\pdfoutput=1
\documentclass[letterpaper]{article}
\usepackage{aaai}
\usepackage{times}
\usepackage{helvet}
\usepackage{courier}
\usepackage{algorithm}
\usepackage{algorithmic}
\usepackage{tabularx}
\usepackage{fixfoot}
\DeclareFixedFootnote{\repnote}{ This is a repated footnote}
\usepackage{multirow}
\begin{document}
%
\title{Structure Learning Using Forced Pruning}
\author{Ahmed Abdelatty\thanks{equal contribution}, Pracheta Sahoo\footnotemark[1], Chiradeep Roy\footnotemark[1]
\\ Department of Computer Science, The University of Texas at Dallas\\
\{Ahmed.Abdelatty, Pracheta.Sahoo, Chiradeep.Roy\}@utdallas.edu\\
}
\maketitle
\begin{abstract}
\begin{quote}
Markov networks are widely used in many Machine Learning  applications including
natural language processing, computer vision, and  bioinformatics .  Learning Markov networks have many complications ranging from intractable computations involved to the possibility of learning a model with a huge number of parameters. In this report, we provide a computationally tractable greedy heuristic for learning Markov networks structure. The proposed heuristic results in a model with a limited predefined number of parameters. We ran our method on 3 fully-observed real datasets, and we observed that our method is doing comparably good to the state of the art methods.
\end{quote}
\end{abstract}

\section{Introduction}
In the past decades, Markov networks (MNs) have been used intensively for developing Machine Learning applications: Computer Vision \cite{pgm-vision,pgm-vision2}, Natural Language processing \cite{lda,pgm-nlp}, Collaborative
Filtering \cite{collab-fill}, and Bioinformatics \cite{Bioinformatics}.

Therefore, many approaches have been proposed to learn both the structure and the parameters of MNs. One approach is to add features greedily in a way that improves the model likelihood \cite{greedy_heuristic,greedy_heuristic1}; which is not allowed to be removed in subsequent iterations.  Such a greedy heuristic can lead to a model with a huge number of features, which will, in turn, leads to overfitting and huge processing time. Alternative methods placed restrictions on the treewidth of the learned structure \cite{Chow_Liu_tree,tree_width,tree_width1}. Such restrictions  on  the  treewidth  guarantees  the tractability of inference  in  the learned model (polynomial time). Finding the optimal bounded treewidth model is NP-hard in general \cite{NP_hard_tree-width} for any bound greater than 1. Another alternative is to use L1-regularization to bias the model towards solutions where many of the parameters are zero \cite{L1_structure,L1_structure1,L1_structure_directed}. However, such models don't give any limit guarantee on the number of the parameters for the learned model.

In this report, we propose a greedy heuristic that limits the number of the parameters for the learned model. The proposed method can be seen as a coordinate ascent optimization, where for the currently fixed structure we update the parameters, following which we then fix the set of parameters and then update the structure greedily.

One approach of MN parameter learning is maximum log-likelihood parameter estimation (MLE). However, in general, MLE is computationally intractable due to the intractability of computing the partition function \cite{Bound-Partition-Function,bethe-partition-function}, alternative tractable approximations use maximum pseudo-log-likelihood estimation (MPLE). Our method uses the penalized MPLE in the parameter learning step (the reason behind using penalization is to prevent overfitting). To be more specific, we used a parameter tying regularization method proposed in \cite{Parameter-Tying}. Parameter tying is the process of  partitioning the parameters into groups and forcing the members of each group to have the same value.

Parameter tying has been proved to be very efficient especially in applications where there are symmetries in the underlying data \cite{Parameter-Tying_sym1,Parameter-Tying_sym2,Parameter-Tying_sym3}. Also, the work introduced in \cite{Parameter-Tying,Parameter-Tying1}, showed that parameter tying can be very efficient when there are no symmetries in the data.

This paper is organized into several sections. The next section deals with structure learning preliminaries that the reader needs to be familiar with in order to understand the approach we followed. The third section explores our algorithm in more detail. The fourth section contains the results we observed while performing our experiments. The fifth section contains potential future work. Finally, we conclude in the sixth section.

\section{Preliminaries}
\subsection{Markov Networks}
A Markov Network or a Markov Random Field is a graphical model that is used to model a joint probability distribution by encoding a subset of its independences. It is an undirected graph $G$ with a set of potential functions $\Phi$. If $X=(X_1,X_2,..,X_n)$ is the collection of random variables in the network, then the joint distribution encoded by a Markov Random Field can be represented by the following equation
\begin{equation}
    P(X=x)=\frac{1}{Z}\prod_{c}\phi_c(x_c)
\end{equation}
where $Z$ is a normalizing constant known as the partition function, $c$ is a clique in $G$ and $x_c$ is the projection of assignment $x$ into all the variables in the clique $c$. In the general case, the scope of the potential functions are usually defined over the maximal cliques. However, there are cases when they are defined over pairwise cliques (edges) only. Such networks are called pairwise Markov Networks. In this paper, we will be limiting our discussion to pairwise networks only.

\subsection{Structure Learning}
There are several existing algorithms for learning the structure of a Markov Random Field from data such as Della Pietra et al.’s algorithm \cite{greedy_heuristic1}. In general, however, most of the state-of-the-art algorithms follow the greedy template below.
\begin{algorithm}
\caption{Greedy Structure Learning}
\label{greedy_structure_learning}
Procedure Greedy-MN-Structure-Search ( \\
\hspace*{0.2cm} $\Omega,$   //all possible features \\
\hspace*{0.2cm} $\mathcal{F}_0,$   //initial set of features \\
\hspace*{0.2cm} score$(\cdot,\mathcal{D}),$   //score \\
)
\begin{algorithmic}[1]
	\STATE $\mathcal{F}' \gets \mathcal{F}_0$ //New feature set
    \STATE $\theta \gets 0$
    \WHILE{termination condition not reached}
    	\STATE $\theta \gets ParameterOptimize(\mathcal{F},\theta,score(\cdot,\mathcal{D}))$ \\
        \STATE // Find parameters that optimize the score objective, \\
        \STATE relative to current feature set, initializing from \\
        \STATE initial parameters
        \FOR{each $f_k \in \mathcal{F}$ such that $\theta_k=0$}
        	\STATE $\mathcal{F} \gets \mathcal{F} - f_k$ \\
            \STATE // remove inactive features
        \ENDFOR
        \FOR{each operator $o$ applicable to $\mathcal{F}$}
        	\STATE Let $\hat{\Delta_o}$ be the approximate improvement for $o$
        \ENDFOR
        \STATE Choose some subset of operators $\mathcal{O}$ based on $\hat{\Delta}$
        \STATE $\mathcal{F}' \gets O(\mathcal{F})$ // apply selected operators to $\mathcal{F}$
    \ENDWHILE
    \RETURN $(\mathcal{F},\theta)$
\end{algorithmic}
\end{algorithm}
Most algorithms use L-1 Regularization in the feature pruning stage because it is a convenient way to eliminate features. In our approach, however, we will be using Automatic Parameter Tying with L-2 Regularization which has been shown to work better than regular learning with L-1 Regularization in most cases \cite{Parameter-Tying}.

\subsection{Automatic Parameter Tying}
Parameter tying is a regularization mechanism that tries to reduce overfitting. In this form of regularization, the parameters are partitioned into clusters and an additional equality constraint is added to all the parameters within a single cluster. In other words, if we have $k$ clusters, then we will have $k$ unique parameters. Usually, the parameters to be tied are mentioned \textit{a priori}, however Chou et al. 2018 uses a novel method to automatically find the $k$ best clusters by minimizing the following objective function
\begin{equation}
\sum\limits_{j=1}^{m}(\theta_j-\mu_{a_j})^{2}
\end{equation}
where $\theta=(\theta_1,..,\theta_m)$ is the collection of original parameters and $\mu_{a_j}$ is the mean of cluster $a$ that parameter $\theta_j$ has been assigned to. We incorporate Automatic Parameter Tying into our approach to prevent overfitting and to further compress our model. For more details, the reader is encouraged to refer to the original paper \cite{Parameter-Tying}.

\subsection{Rejection Sampling}
Rejection sampling is a technique to sample from an easy distribution $q(x)$ instead of a difficult distribution $p(x)$. More formally, a sample $s$ is generated from $q(x)$ followed by a number $u$ that is generated from the uniform distribution $U(0,1)$. Sample $s$ is accepted iff the following condition holds
\begin{equation}
u \le \frac{p(x)}{Mg(x)}
\end{equation}
where M is a large constant such that $Mg(x)$ completely envelops $p(x)$ i.e. $Mg(x) \ge p(x)$ for all $x$. In our approach, we use rejection sampling as one of the two options for deleting edges from our current model.

\begin{table*}[ht]
\centering
\begin{tabularx}{\linewidth}{|l|X|X|X|X|X|X|X|X|X|X|X|X|X|X|X|}
	\hline
    \multirow{2}*{Datasets} & \multicolumn{3}{|c|}{m=0}
    		& \multicolumn{3}{|c|}{m=15} & \multicolumn{3}{|c|}{m=30} 
            & \multicolumn{3}{|c|}{m=45} & \multicolumn{3}{|c|}{m=60} \\
    \cline{2-16}
    & k=0 & k=5 & k=10 &
    k=0 & k=5 & k=10 &
    k=0 & k=5 & k=10 &
    k=0 & k=5 & k=10 &
    k=0 & k=5 & k=10 \\
    \hline
    nltcs 
    	&5.08 &4.95 &4.96
        &4.75 &4.71 &4.71
        &4.75 &4.71 &4.71
        &4.77 &4.71 &4.71
        &4.75 &4.73 &4.72 \\
    plants
    	&12.25 &11.43 &11.15
        &12.22 &11.40 &11.11
        &12.37 &11.21 &11.20
        &12.27 &11.25 &11.15
        &12.25 &11.23 &11.17 \\
    msnbc
    	&6.29 &6.15 &6.06
        &5.71 &5.59 &5.52
        &5.71 &5.57 &5.53
        &5.72 &5.57 &5.53
        &5.71 &5.57 &5.53 \\
    \hline
\end{tabularx}
\caption{Negative PLL scores for Greedy Edge Deletion}
\label{table:1}
\end{table*}

\begin{table*}[ht]
\centering
\begin{tabularx}{\linewidth}{|l|X|X|X|X|X|X|X|X|X|X|X|X|X|X|X|}
	\hline
    \multirow{2}*{Datasets} & \multicolumn{3}{|c|}{m=0}
    		& \multicolumn{3}{|c|}{m=15} & \multicolumn{3}{|c|}{m=30} 
            & \multicolumn{3}{|c|}{m=45} & \multicolumn{3}{|c|}{m=60} \\
    \cline{2-16}
    & k=0 & k=5 & k=10 &
    k=0 & k=5 & k=10 &
    k=0 & k=5 & k=10 &
    k=0 & k=5 & k=10 &
    k=0 & k=5 & k=10 \\
    \hline
    nltcs 
    	&5.89 &5.88 &5.88
        &5.59 &5.58 &5.58
        &5.51 &5.44 &5.46
        &5.33 &5.14 &5.13
        &5.38 &4.88 &4.90 \\
    plants
    	&12.91 &13.36 &11.15
        &12.82 &13.29 &13.29
        &12.79 &13.30 &13.28
        &12.91 &13.31 &13.29
        &13.01 &12.92 &13.16 \\
    msnbc
    	&6.40 &6.55 &6.63
        &5.83 &5.83 &5.92
        &5.77 &5.82 &5.86
        &5.75 &5.77 &5.77
        &5.77 &5.69 &5.77 \\
    \hline
\end{tabularx}
\caption{Negative PLL scores for Rejection Sampling}
\label{table:2}
\end{table*}

\section{Our Approach}
In our approach, we want to explore all models having exactly $M$ edges (that might be reduced further using parameter tying in order to reduce overfitting). We want to start out with an initial model having $M$ edges and then keep deleting and adding $k$ edges from the remaining pool of edges. During the addition step, we greedily choose the edges based on their individual contribution to the PLL score of the current model. Similarly, during the deletion step we remove the $k$ \textit{worst} edges based on their individual contributions to the PLL score. However, while this is a reasonable heuristic, there might be cases where we keep exchanging the same $k$ edges over and over again. In addition, the globally optimal group of $k$ edges might not individually be the top contributors to the PLL score.

Ideally, we want to choose the edges to delete from the distribution $P_d$ over all $k$ groups of edges in the current model. For the greedy approach, we would probably choose the MAP assignment; however, as we mentioned earlier, there might be cases where the same edges keep getting deleted and replaced multiple times. In order to circumvent this problem, we would ideally want to sample $k$ edges from $P_d$ instead. We do that in our algorithm by using rejection sampling.

Given below is our approach according to the generic greedy template. Since we want to compare the sampling approach with the individual greedy deletion approach, we will include the heuristic as a parameter to our algorithm.

\begin{algorithm}
\caption{Forced Pruning Learner}
\label{forced_pruning_algorithm}
 \textbf{Input:} \hspace*{0.15cm} Training dataset \textit{d1}, Testing dataset \textit{d2}, \\
 \hspace*{1.22cm} Model size \textit{M}, Exchange size \textit{k}, Heuristic \textit{h}\\
 \textbf{Output:} Model of size \textit{M}
\begin{algorithmic}[1]
            \STATE $C \gets GenerateCompleteModel(d1)$
            \STATE $T \gets ChowLiuTree(d1)$
            \STATE $P \gets ExtractPotentials(T)$
            \STATE $R \gets ExtractPotentials(C) - P$
    		\STATE $P \gets P \cup ExtractRandom(M-|P|,R)$
            \FOR{$i \in \{1..maxiter\}$}
            	\STATE $\theta \gets LearnParamsWithAPT(P)$
                \STATE $P_d \gets DeletePotentials(P,\theta,h,k)$
                \STATE $P_a \gets AddPotentials(P,\theta,R,k)$
                \STATE $P \gets (P - P_d) \cup P_a$
                \STATE $R \gets (R - P_a) \cup P_d$
            \ENDFOR
            \RETURN P
        \end{algorithmic}
\end{algorithm}

The $DeleteEdges(P,\theta,h,k)$ will remove $k$ edges from the potential structure P given parameters $\theta$ according to heuristic $h$. In this report, we will be using both greedy deletion as well as rejection sampling. The $AddPotentials(P,\theta,R,k)$ will choose $k$ potentials from $R$ that result in the highest PLL gain. The rejection sampling procedure is defined on the distribution on all possible $k$-vectors of edges that can be deleted. The distribution is proportional to the likelihood of the model, given the deletion of the $k$-vector in question so we set $Mg(x)$ to be $1.0$ since that is the largest possible likelihood of all possible models and is therefore an upper bound.

\section{Experiments}
We evaluated Forced Pruning on 3 real-world datasets - nltcs, plants and nsbc. We compared the difference in results for both the greedy deletion as well as the rejection sampling heuristics. For each dataset we experimented by adding $m={0,15,30,45,60}$ edges to the original Chow-Liu tree and tried running the algorithm by switching $k={1,5,10}$ edges at every iteration.

We noticed that the greedy heuristic outperformed the rejection sampling heuristic in almost all the cases. A possible cause for this could be the fact that in the case of using greedy edge deletion, we are approaching a local optimum with the best possible set of $M$ edges. In the rejection sampling experiments, there is a certain element of randomness that keeps jumping around the function. Increasing the number of $k$ edges to be switched out in the greedy case almost always leads to a better PLL score as k is increased. In the rejection sampling experiments however, the PLL scores sometimes get worse on increasing $k$. A possible reason for this is the fact that there is more randomness in choosing a larger $k$-vector than a smaller one. Overall, there is a general trend towards the PLL scores improving as the size of the model increases, which is a good sign.

\section{Future Work}
One of the major bottlenecks we faced while running our experiments was the fact that the ones involving rejection sampling took a lot of time to complete because of a large number of samples getting rejected at every iteration of the algorithm (especially if the initial model has a bad likelihood score). A future version of the algorithm can use a form of MCMC sampling to do the sampling. In addition, we can also try sampling from the distribution on the $k$ edges to be added to the model after deleting $k$ edges from it. However, the results in our experiments show that individual greedy deletion appears to fare better than the sampling approach and has also much lower time complexity thereby making the edge deletion procedure cheaper. In the future, we plan on modifying the algorithm to do rejection or MCMC sampling with some amount of randomness or when a local optimum is reached. This approach, we hope, shall improve the PLL scores in the final model trained using the new algorithm.

\section{Conclusion}
In this paper, we proposed an algorithm to explore pairwise Markov Networks of size $M$ and find the best possible such model by switching $k$ edges in and out of the model. We used two heuristics to accomplish this, namely using individual greedy edge deletion and rejection sampling from the distribution on all possible $k$-vectors. We ran a number of experiments on three datasets and concluded that the greedy heuristic performs better than sampling in most cases. We also proposed future improvements to the algorithm such as using MCMC instead of rejection sampling, studying the effect of sampling during addition of edges and introducing some form of random restarts in order to avoid local optima.

\bibliographystyle{plainnat}

\end{document}